\documentclass[letterpaper, 10 pt, journal, twoside]{IEEEtran}
\ifCLASSINFOpdf
\else
\fi
\usepackage{xcolor}
\usepackage{graphicx}
\graphicspath{{Figs/}} 
\usepackage{amssymb,amsmath}
\DeclareMathOperator{\sgn}{sgn}
\usepackage{multirow}
\usepackage{array}
\newcolumntype{P}[1]{>{\centering\arraybackslash}p{#1}}
\usepackage{pgfplots} 
\pgfplotsset{compat=1.15,
    dottedstyle/.style={
        rdot/.style={red,mark=none, thick, dotted},
        bdot/.style={blue,mark=none, thick, dotted},
        gdot/.style={green,mark=none, thick, dotted},
        rlin/.style={red,  mark=none, very thick },
        blin/.style={blue, mark=none, very thick },
        glin/.style={green,mark=none, very thick },
    },
}
\usepackage{wasysym}

\usepackage{hyperref}

\definecolor{red}{rgb}{1.00,0.00,0.00}
\definecolor{lightred}{rgb}{1.00,0.3,0.3}
\definecolor{blue}{rgb}{0.00,0.00,1.00}
\definecolor{green}{rgb}{0.1,0.50,0.1}
\definecolor{yellow}{rgb}{0.5,0.5,0.0}
\definecolor{white}{rgb}{1,1,1}
\definecolor{gray}{rgb}{0.4,0.4,0.4}

\begin{document}
%
\title{A CPG-Based Agile and Versatile Locomotion Framework with Proximal Symmetry Loss Function}
%
%
%

\author{Mohammadreza Kasaei$^{1}$, Miguel Abreu$^{2}$, Nuno Lau$^{1}$, Artur Pereira$^{1}$, and Lu\'is Paulo Reis$^{2}$%
\thanks{$^{1}$Mohammadreza Kasaei, Nuno Lau and Artur Pereira are with IEETA/DETI, University of Aveiro, Portugal
        {\tt\footnotesize \{mohammadreza, nunolau, artur\}@ua.pt}}%
\thanks{$^{2} $Miguel Abreu and  Lu\'is Paulo Reis are with LIACC/FEUP, Artificial Intelligence and Computer Science Lab, University of Porto, Portugal
        {\tt\footnotesize \{m.abreu,lpreis\}@fe.up.pt}}%
\thanks{Digital Object Identifier (DOI): see top of this page.}
}
%
%

\markboth{}
{Kasaei \MakeLowercase{\textit{et al.}}: A CPG-Based Agile and Versatile Locomotion Framework} 

%



\maketitle

\begin{abstract}
Humanoid robots are made to resemble humans but their locomotion abilities are far from ours in terms of agility and versatility. When humans walk on complex terrains, or face external disturbances, they combine a set of strategies, unconsciously and efficiently, to regain stability. This paper tackles the problem of developing a robust omnidirectional walking framework, which is able to generate versatile and agile locomotion on complex terrains. The Linear Inverted Pendulum Model and Central Pattern Generator concepts are used to develop a closed-loop walk engine, which is then combined with a reinforcement learning module. This module learns to regulate the walk engine parameters adaptively, and generates residuals to adjust the robot's target joint positions (residual physics). Additionally, we propose a proximal symmetry loss function to increase the sample efficiency of the Proximal Policy Optimization algorithm, by leveraging model symmetries and the trust region concept. The effectiveness of the proposed framework was demonstrated and evaluated across a set of challenging simulation scenarios. The robot was able to generalize what it learned in unforeseen circumstances, displaying human-like locomotion skills, even in the presence of noise and external pushes.

\end{abstract}


%
\IEEEpeerreviewmaketitle

\section{Introduction}\label{sec:introduction}
%
%
%
%
\IEEEPARstart{V}{ersatility} is a coveted feature when designing humanoid robots. Their shape allows them to be extremely resourceful in our daily-life environments without the necessity of adjusting the surroundings. According to this distinctive property, humanoid robots have a wide range of applications, from working in factories to helping elderly people. Despite a significant effort from the research community, their capabilities are still far from ours, particularly, in terms of speed, stability and safety. People expect humanoid robots to walk robustly over any type of terrain and be able to recover from external perturbations.

Many researchers have conducted studies to find solutions for this problem. These studies deal mostly with developing robust walking frameworks and can be categorized into four major categories:~\textit{model-based}, \textit{model-free}, \textit{Central Pattern Generator~(CPG)-based} and \textit{hybrid} frameworks. \textit{Model-based} frameworks have been designed by considering the physical dynamics model of the systems, but uncertainties often affect their performance and prevent them from achieving an optimal performance. \textit{Model-free} frameworks are based on machine learning~(ML) algorithms like reinforcement learning~(RL). These approaches are trial intensive and need a considerable amount of samples that cause many falls. Consequently, they cannot be used directly on real robots. \textit{CPG-based} frameworks have been inspired by nature and designed based on neuro-physiological studies on animals. In these approaches, a set of oscillators are coupled together in a specific manner to generate rhythmic locomotion. Their greatest strengths are their flexibility, adaptability and computational efficiency. On the other side, the difficulties in adjusting the parameters of the oscillator and adapting sensory information are their worst weaknesses. \textit{Hybrid} frameworks combine the aforementioned approaches to leverage their different capabilities.

In this paper, we propose an omnidirectional hybrid walking framework that is able to generate versatile and agile locomotion on complex terrains, even in the presence of noise and external pushes. Specifically, we use the Linear Inverted Pendulum Model~(LIPM)~\cite{kajita20013d} concept to couple a set of Partial Fourier Series~(PFS) oscillators in order to design a closed-loop CPG-ZMP-based walk engine. This engine is then combined with an extension of the Proximal Policy Optimization (PPO) algorithm~\cite{schulman2017ppo} that leverages model symmetries to learn model-free skills and tune the parameters of the walk engine adaptively. We apply this approach to the COMAN~\cite{tsagarakis2011coman} humanoid robot in order to learn residual physics --- a term originally used by Zeng et al.~\cite{zeng2020tossingbot}. In this strategy, residuals are the difference between the parameters computed by the analytical walk engine and the optimal parameters, which we seek to estimate through reinforcement learning.

One of the most recognizable features of humans and bilateral animals is their approximate symmetry with respect to the sagittal plane. Humanoid robots, as the name implies, share some fundamental biomechanical characteristics with humans, including symmetry, which is usually assumed to be exact during simulation. In general, model minimization is conceptually straightforward when dealing with symmetries. However, in practice, there are several techniques to leverage this property in order to increase the sample efficiency or improve the learned behavior. Most of them fall under spatial symmetry and, to a lesser extent, temporal symmetry. The latter concerns invariability under temporal transformation, such as scaling or inversion. On the other side, techniques based on spatial symmetry are more common, and can be split into diverse areas, as described in the following paragraphs. 

\subsection{Relabeling states and actions}

The first spatial symmetry technique consists in relabeling states and actions, allowing the permutation of roles between elements of the same type. As an example, consider a simple robot with 3 rigid body parts (a fixed torso and two arms) that has to hit a ball that is thrown into the air. The state of the robot is composed of three variables, indicating the relative position of each leg and the ball, in relation to the torso. An action is composed of two variables, to control the position of each arm. If the robot learns how to hit the ball with the right arm, we can easily transpose that knowledge to the left arm by switching both arms in the state and actions spaces of the robot (without forgetting to mirror all relative positions, horizontally). This is a low-level approach that can be implemented directly on the simulator and requires no changes to the optimization algorithm. 

\subsection{Data augmentation}

Data augmentation is widely used in RL to improve sample efficiency and stability, being one of its main precursors the experience replay method introduced by Lin~\cite{lin1992}. In the scope of this work, data augmentation consists in artificially creating samples by applying a symmetric transformation to actually experienced samples. Using the last example, consider that, for 5 seconds, the robot was waving with the right arm and poking a ball with the left arm. As far as the RL algorithm knows, the robot also spent 5 seconds doing the symmetric action (where the roles of both arms were switched). In the end, there are 10 seconds of useful experience that can be used for learning. In theory, the algorithm learns the same actions with both arms, and there is no dominant arm. In practice, this is not guaranteed in typical implementations of RL optimization algorithms due to the asymmetric bias introduced in the beginning of the optimization, when initializing the parameters of the policy.

\subsection{Symmetric Networks}

Symmetric networks comprise the third subdivision of spatial symmetry. These solutions enforce symmetry constraints directly on the policy, by modifying the network architecture. This approach is very robust in applications where symmetry is an intrinsic characteristic of optimal solutions. For example, given an optimal ball throwing technique with the right hand and assuming perfect model symmetry, we can automatically infer the optimal technique for the left hand by applying a reflection transformation. However, this equivalence cannot be assumed for a cyclic task, like biped walking, where the optimality of each step cannot be assessed individually. It is therefore unclear whether a perfectly symmetrical gait is preferable or not. Additionally, unbalanced gait patterns (within reason) can be perceived as unimpaired or normal \cite{handvzic15}.

\subsection{Symmetry Loss Function}

In an optimization problem, a loss function represents the cost associated with an event. The objective is to find the best solution by minimizing that cost. As an example, if a robot is learning how to maintain its equilibrium, the more times it falls, the higher the cost. Analogously, when learning symmetric tasks, the degree of asymmetry in the robot's actions can also be seen as a cost. In a reinforcement learning problem, this strategy may be confused with reward shaping. However, from the perspective of the optimization algorithm, there is a crucial difference. The reward is a simple scalar value, whereas the cost function can be used to steer the optimization in the right direction, by yielding a gradient (which indicates the direction in which the algorithm's parameters should be changed). A loss function constitutes a flexible way of incorporating symmetry in an optimization problem. It allows more freedom at both design time and runtime than any of the other methods. However, it gives no guarantee of symmetry, like the relabeling technique or symmetric networks.

Concerning problems which do not necessarily benefit from perfect symmetry, the symmetry loss function is the most advantageous approach because it allows the user to define exactly how important symmetry is in the policy, at any given moment during learning. Moreover, it does not get stuck in neutral states (see Section~\ref{sec:relatedwork_symnet}) and it is more computationally efficient than data augmentation, especially when dealing with large sample batches. However, powerful policy optimization algorithms such as PPO~\cite{schulman2017ppo} cannot take full advantage of existing solutions. To close this gap, we propose a new loss function called Proximal Symmetry Loss, which attempts to leverage the trust region concept that characterizes the Proximal Policy Optimization algorithm.

Our contribution is twofold. First, the development of a hybrid walking framework that connects a closed-loop \mbox{CPG-ZMP} walk engine with a reinforcement learning module that learns residual physics. Secondly, the improvement of sample efficiency and natural looking behaviors, by extending PPO with a novel symmetry loss function.

\section{Preliminaries}

The problem of learning on top of a CPG-ZMP walk engine can be described as a Markov Decision Process (MDP) -- a tuple $\left\langle \mathcal{S,A},\Psi,p,r\right\rangle$, with a set of states $\mathcal{S}$, a set of actions $\mathcal{A}$, a set of possible state-action pairs $\Psi\subseteq \mathcal{S}\times \mathcal{A}$, a transition function $p(s,a,s'):\Psi \times \mathcal{S} \rightarrow[0,1]$, and a reward function $r(s,a):\Psi \rightarrow {\rm I\!R}$.

Model reduction allows the exploitation of redundant or symmetric features. To this end, Ravindran and Barto~\cite{ravindran2001} proposed a mathematical formalism to describe MDP homomorphisms --- a transformation that groups equivalent states and actions. An MDP homomorphism $h$ from $M=\left\langle \mathcal{S,A},\Psi,p,r\right\rangle$ to $\bar{M}=\left\langle \mathcal{\bar{S},\bar{A}},\bar{\Psi},\bar{p},\bar{r}\right\rangle$ can be defined as a surjection $h:\Psi \rightarrow \bar{\Psi}$, which is itself defined by a tuple of surjections $\left\langle f,\{g_s|s\in \mathcal{S}\} \right\rangle$. In other words, equivalent state-action pairs in $M$ are mapped by $h$ to the same abstract state-action pair in $\bar{M}$. For $(s,a)\in \Psi$, the surjective function $h((s,a))=(f(s),g_s(a))$, where $f:\mathcal{S}\rightarrow \mathcal{\bar{S}}$ and $g_s:\mathcal{A}_s\rightarrow \mathcal{\bar{A}}_{f(s)}$ for $s\in \mathcal{S}$, satisfies:

\begin{align}
\bar{p}(f(s),g_s(a),f(s')) & = \sum_{s''\in [s']_B}p(s,a,s''), \label{eq:homTran} \\ 
& \quad\; \forall s, s' \in \mathcal{S},a\in \mathcal{A}_s,  \nonumber
\\
\mathrm{and} \quad  \bar{r}(f(s),g_s(a)) & =r(s,a), \quad \forall s \in \mathcal{S}, a \in \mathcal{A}_s,  \label{eq:homRew}
\end{align}


\noindent
where $B$ is a partition of $\mathcal{S}$ into equivalence classes, and $[s']_B$ denotes the block of partition $B$ to which state $s'$ belongs.

MDP symmetries constitute a specialization of the described framework, where $f$ and $g_s, s\in \mathcal{S}$ are bijective functions and, consequently, the homomorphism $h=\left\langle f,\{g_s|s\in \mathcal{S}\} \right\rangle$ from $M$ to $\bar{M}$ is an isomorphism. Additionally, since symmetries can be characterized as MDP isomorphisms from and to the same MDP, they are automorphisms, which simplifies the homomorphism conditions~\eqref{eq:homTran} and~\eqref{eq:homRew}:

\begin{align}
p(f(s),g_s(a),f(s')) & = p(s,a,s'), \quad \forall s, s' \in \mathcal{S},a\in \mathcal{A}_s, \label{eq:homTran2}\\
\mathrm{and} \quad  r(f(s),g_s(a)) & =r(s,a), \quad \forall s \in \mathcal{S}, a \in \mathcal{A}_s. \label{eq:homRew2}
\end{align}

This means that symmetries can be defined in a single MDP, $M$. As an example, consider a symmetric humanoid robot learning how to walk forward. If the robot is leaning to the left side (state $s$), it may attempt to move its left foot (action $a$) so that the projection of its center of mass lies inside its base of support, ending in a stable state $s'$. In this case, $f(s)$ would correspond to leaning to the right side, $g_s(a)$ would be moving the right foot, and $f(s')$ would be a stable state symmetric to $s'$ with respect to the robot's sagittal plane. Note that it is possible that $s'=f(s')$. Equation \eqref{eq:homTran2} holds in this scenario because the probability of ending in a stable state by performing action $a$ in $s$ is the same as performing $g_s(a)$ in $f(s')$. As for \eqref{eq:homRew2}, it also holds, since the reward in both cases should be the same, otherwise it would contain an asymmetry bias.

\section{Related Work}
Several solutions have been proposed in the literature to combine a model-based or a CPG-based walk engine with machine learning~(ML) algorithms to generate robust locomotion. This section starts by presenting specific techniques to leverage symmetries in MDPs, followed by a review of some recent works concerning the development of hybrid walking frameworks.

\subsection{Symmetries in MDPs} \label{sec:relatedwork_sym}

Following the overview about symmetries in MDPs presented in Section~\ref{sec:introduction}, we delve now into specific techniques, as well as their strengths and weaknesses. In the context of temporal symmetry, time inversion requires the application to be characterized as a conservative system with no loss of mechanical energy (e.g. due to friction), such as a frictionless pendulum, which Agostini and Celaya \cite{agostini2009exploiting} control through RL. Since this assumption is unrealistic in real world systems, this approach is not desirable. Regarding spatial symmetry, as aforementioned, the contributions can be majorly subdivided into:

\subsubsection{Relabeling states and actions} Surovik et al. \cite{surovik2019adaptive} combine this technique with reflection transformations to swap frame-dependent values, such as lateral rotation, to effectively reduce the state volume of a tensegrity robot. However, this same idea can be applied to biped models in an intuitive way. If, instead of left and right, we think of "stance" and "non-stance" leg, we can relabel the physical legs every half cycle to obtain a symmetric controller~\cite{xie2020learning,hereid2018dynamic}. Peng et al.~\cite{peng2017deeploco} apply this relabeling method at fixed intervals of 0.5~s, forcing the gait cycle to have a fixed period. This idea is simple to implement but it restricts the gait dynamics, hindering fast recoveries from unexpected external forces. In all cases, the policy is constrained by some analytical symmetry switch, whether it is based on time or behavioral pattern.


\subsubsection{Data augmentation} In symmetry oriented solutions, data augmentation can be used with model-based~\cite{bree2021data} or model-free RL algorithms, although the scope of this work is limited to the latter alternative. Examples of successful applications of this technique include a real dual-armed humanoid robot that moves objects~\cite{lin2020}, the walking gait of several humanoid models~\cite{abdolhosseini2019learning} and a quadruped with more than one plane of symmetry~\cite{mishra2019augmenting}, among others~\cite{agostini2009exploiting,silver2016go}. This approach is computationally inefficient for large sample batches. 

\subsubsection{Symmetric Network} \label{sec:relatedwork_symnet} Concerning humanoid models, Abdolhosseini et al. \cite{abdolhosseini2019learning} introduced a symmetric network architecture that forces perfect symmetry. This method guarantees that if states and actions are symmetrically normalized, the behavior has no asymmetrical bias. However, as the authors acknowledge, neutral states are inescapable, i.e., a symmetrical policy $\pi(s)=g_{f(s)}(\pi(f(s))), \forall s \in \mathcal{S}$ applied to a symmetrical state $s=f(s)$ cannot yield asymmetrical states unless the environment introduces its own bias. Pol et al.~\cite{Pol2020mdp} generalize this approach to additional problems by introducing MDP homomorphic networks, which can be automatically constructed by stacking equivariant layers through a numerical algorithm. The equivariance constraints (under a group of reflections or rotations) are applied to the policy and value networks. This family of solutions lacks the ability to control the symmetry enforcement at runtime and the ability to symmetrize asymmetrical models.

\subsubsection{Symmetry Loss function} Mahajan and Tulabandhula~\cite{mahajan2017symmetry} proposed an automated symmetry detection process and a method of incorporating that knowledge in the Q-learning algorithm \cite{watkins1992q}. A soft constraint was presented as an additional loss for the Q-function approximator, $\hat{\mathbb{E}}_t[(Q_\theta(f(s_t),g_{s_t}(a_t))-Q_\theta(s,a))^2]$, where $\theta$ is the parameter vector of $Q$, and the expectation $\hat{\mathbb{E}}_t$ indicates the empirical average over a batch of samples. Yu et al. \cite{yu2018learning} transpose this idea to policy gradients by fusing a new curriculum learning method with an homologous loss function, $\sum_i\|(\pi_\theta(s_i)-g_{f(s_i)}(\pi_\theta(f(s_i)))\|^2$. The authors present very good results when both approaches are used simultaneously. However, without the learning curriculum, the symmetry loss has no significant impact in the learned policy.

\subsection{Combination of Model-Based and ML}

Yang et al.~\cite{yang2018learning} proposed a stabilizer framework based on deep reinforcement learning (DRL) to acquire motor skills for push recovery in humanoid robots. Their framework is composed of two layers that are connected hierarchically. The high-level layer generates target joint angles and the low-level layer is responsible for tracking these angles using a set of PD controllers. In their approach, a high-level planner and feedback control were integrated as a single coherent policy network by designing a specific reward function based on traditional push recovery strategies. The reward function was composed of six distinctive terms that are added together with different weights. They conducted a set of simulations using a simulated NASA Valkyrie robot to validate the performance of their approach. The results showed that the robot learns to deal with different types of disturbances to regain its stability.

Koryakovskiy et al.~\cite{koryakovskiy2018model} argued that learning from scratch approaches are not applicable on real robots due to involving many trials with exploratory actions. To circumvent this problem, they combined RL with a model-based control using two different approaches to compensate model-plant mismatches. The first approach consisted in learning a compensatory control action that minimizes the same performance measure as a nominal controller. The second approach was designed to learn a compensatory signal from the difference between an actual transition and a predicted transition by the internal model~(MPML). Using a set of simulations, they showed that MPML is not only twice as fast, but it also exhibits no deviating behaviors. Moreover, they tested MPML using a real robot named Leo in a squat scenario. The results showed that MPML successfully realizes squatting by learning the compensation signal.

Ahn et al.~\cite{ahn2020data} formulated humanoid locomotion as a novel Markov Decision Process~(MDP) for safe and data-efficient learning, aided by a dynamic balancing model. Specifically, they proposed a structured footstep planner by combining a time-to-velocity-reversal planner based on LIPM with a neural network and a safety mechanism. In their structure, the planner generates achievable sub-optimal guidance, and then the neural network is used to maximize the long-term reward. The safety controller takes into account the capturability concept to facilitate safe exploration. To evaluate the performance of their method, they carried out a series of simulations using two different simulated robots. The results showed that their method can generate various types of walking, including turning and walking on irregular terrains robustly. 

\subsection{Combination of CPG-Based and ML}

Tran et al.~\cite{tran2018humanoid} proposed a perturbation rejection based on reinforcement learning on top of a CPG walking generator. Specifically, they combined a self-organizing map with a Q-learning algorithm to automatically detect and select appropriate reactions to regain the stability during the swing phase. The effectiveness of their approach has been validated by simulations and real robot experiments using a NAO robot. The results showed that their approach was only able to reject perturbations when the robot was performing stable repetitive movements before being pushed.

Wang et al.~\cite{wang2018matsuoka} developed a framework for adaptive walking of humanoid robots based on Matsuoka’s CPG. Their framework was composed of three main stages including feature processing, neural network (NN) learning, and signal filtering. Particularly, the first stage transforms Matsuoka’s CPG outputs into a normalized limit cycle which will be combined with sensory feedback using a three-layer NN to generate a set of rhythmic signals for walking. In the last stage, they used a first-order low-pass filter to avoid discontinuity in the output of the NN. The performance of their method has been validated through a set of simulations and experiments using a NAO robot. The results showed that their framework was able to generate adaptive walking on fixed and inclined terrains based on sensory feedback. 

Van der Noot et al.~\cite{van2018bio} developed a bio-inspired controller for biped locomotion. Their controller utilizes virtual muscles driven by reflexes and a central pattern generator to produce human-like and energy efficient locomotion. In their structure, the controller generates torque references for all the joints based on the biped's state. Additionally, the user can provide high-level commands as linear or quadratic functions. They deployed their framework on a simulated COMAN and showed a robust and human-like walking.

The combination of CPG (or model-based) and ML is not limited to biped robots. This approach has been applied to many types of robots, from quadrupeds~\cite{lee2020learning,saputraneural} to snakes~\cite{snake} to generate bio-inspired and robust locomotion for traversing complex environments.

\begin{figure}[!t]
    \centering
    \includegraphics[width = 1.0\columnwidth]{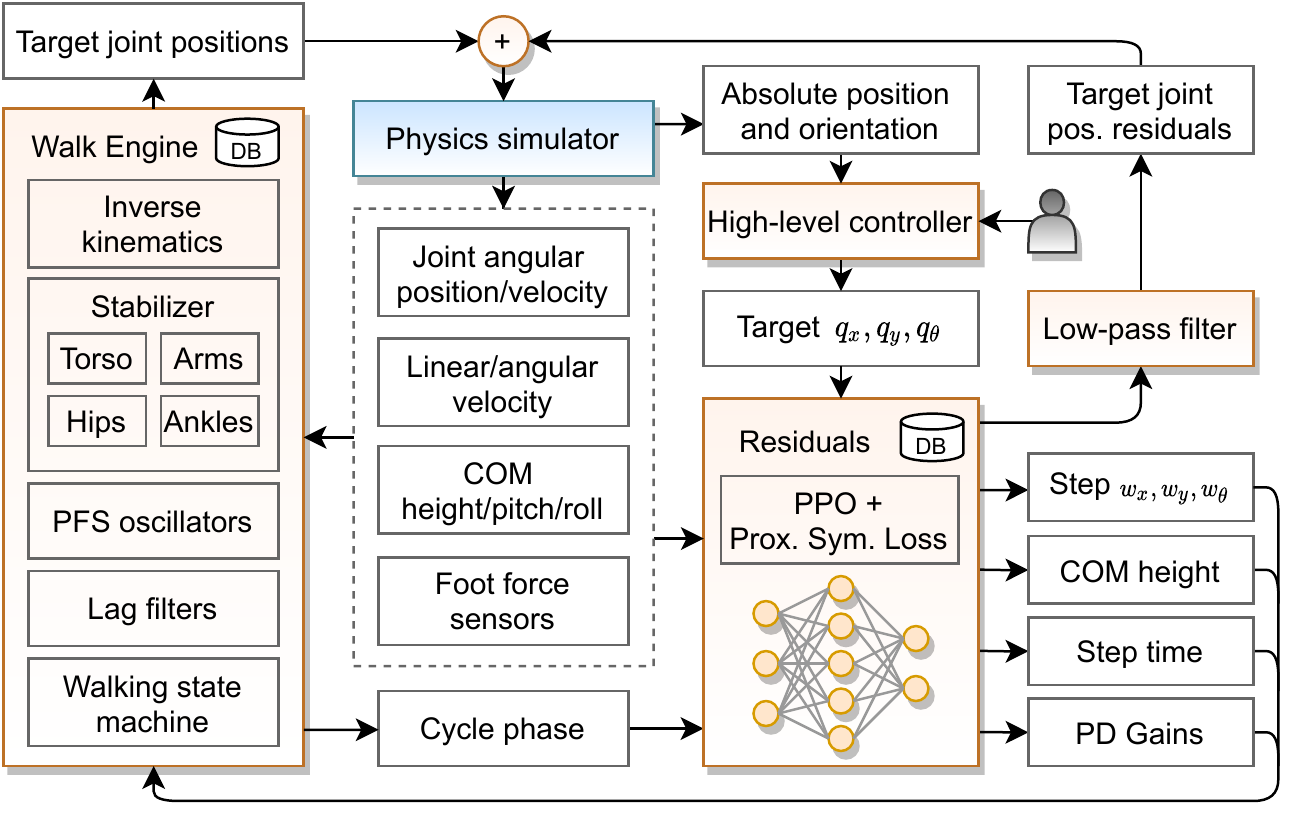}
	\caption{Overall architecture of the proposed framework. The highlighted boxes represent the main modules and the white boxes are exchange information among them.}
	\label{fig:arch}
\end{figure}

\section{Architecture}

An overview of the proposed architecture is depicted in Fig.~\ref{fig:arch}. The highlighted boxes correspond to functional modules, while the white boxes represent data that is exchanged between them. This framework is composed of three main components: the \texttt{physics simulator}, the \texttt{walk engine} and the \texttt{residuals}. The employed simulator was PyBullet~\cite{coumans2020}, which is based on the open-source Bullet Physics Engine. PyBullet is responsible for simulating the interaction of the COMAN robot with the environment and produce information about the robot's joints and sensors for the \texttt{walk engine} and \texttt{residuals}, as well as its position and orientation for the high-level controller. Based on this information, the high-level controller generates a new target position and orientation, according to the restrictions of the user and the current scenario, and outputs $q_x,q_y$ and $q_\theta$ in the robot's local reference frame. Then, the \texttt{residuals} module uses those values along with the robot's observations and the cycle phase (which is initialized to zero), and generates parameters for the \texttt{walk engine}, including control commands for the step size $w_x,w_y$ and step rotation $w_\theta$, the target center of mass (COM) height, the duration of a step, and PD gains. At the same time it also outputs target joint position residuals, which go through a low-pass filter.

Finally, the \texttt{walk engine} generates a set of trajectories based on CPG and analytical control approaches according to the parameters dictated by the \texttt{residuals} module, and the robot's observations. Then, the target joint positions are calculated by an inverse kinematics method and added to the position residuals. The result is fed to the physics simulator, which advances one time step. During the reinforcement learning stage, the \texttt{residuals} module is optimized using PPO and a proximal symmetry loss. In the testing phase, a static snapshot of the best policy is employed instead.

\section{Walk Engine} \label{sec:walk_engine}
In this section, the problem of generating a walking gait will be addressed by presenting a modular walk engine based on the combination of LIPM with CPG. In this walk engine, a state machine is designed to control the overall gait and to generate walking reference trajectories. These are obtained by arranging a set of Partial Fourier Series~(PFS) oscillators and combining them with a set of controllers to stabilize the robot. 

\subsection{Walking State Machine}
Due to the symmetric and periodic nature of humanoid locomotion, it can be represented by a state machine which is composed of five distinct states: \textit{Idle}, \textit{Init Single Support}, \textit{Single Support}, \textit{Init Double Support}, and \textit{Double Support}. In this state machine, a timer is assigned to each state and, normally, a transition will occur once the timer is over. Moreover, in some situations, such as when the robot should immediately place its foot down, the state transition can be issued by specific controllers. In the \textit{Idle} state, the robot is standing and waiting for a new walking command \mbox{($\boldsymbol{W} = [w_x, w_y, w_z, w_\theta, w_t]^\top$)} which contains length,  width, height, rotation and duration of the next stride, respectively.
The system transits to \textit{Init Single Support} once the new command is received. It should be noted that the new command is always passed through a first-order lag filter to ensure a smooth update. In this state, all the reference trajectories will be generated based on the filtered command. Then, the robot shifts its COM towards the first support foot to be ready to lift its swing foot. Afterward, the system will automatically transit to \textit{Single Support} and the robot tracks the generated references to move its swing leg towards the next support foot position. Then, in \textit{Init Double Support}, both feet are adjusted to be in contact with the floor. Finally, during \textit{Double Support}, the robot shifts its COM towards the next support foot and, at the end of this state, the state will transition to \textit{Init Single Support} if another step is needed, or \textit{Idle}, otherwise. It should be noted that the new command is applied just at the beginning of \textit{Init Single Support}.

\subsection{Walking Reference Trajectories}
To generate walking reference trajectories, a set of footsteps ($\boldsymbol{f}_i = [f_{i_x} \quad f_{i_y} ]^\top, \; i \in \mathbb{N} $) are computed according to the input stride's parameters~($\boldsymbol{W}$) and the current feet configuration. Then, the LIPM concept is used to model the overall dynamics of a humanoid robot using a differential equation:
\begin{equation}
    \ddot{\boldsymbol{c}} = \omega_0^2 (\boldsymbol{c} - \boldsymbol{p}) ,
\label{eq:LIPM}    
\end{equation}
\noindent
where \mbox{$\boldsymbol{c}= [c_x\quad c_y]^\top$} denotes the position of the COM, \mbox{$\boldsymbol{p} = [p_x\quad p_y]^\top $} is the position of the ZMP and \mbox{$\omega_0 = \sqrt{\frac{g}{c_z}}$} represents the  natural frequency of the pendulum, where $g$ and $c_z$ are the gravity constant and the vertical position of the COM, respectively. 

According to the LIPM, motion equations in sagittal and frontal planes are equal and independent. Additionally, since the walking gait is symmetric, the reference trajectories can be generated just for one side and then applied to both sides with a half-period phase shift. Accordingly, a set of PFS oscillators are arranged to generate walking reference trajectories based on LIPM in the sagittal plane. A PFS oscillator can be modeled as follows:
\begin{equation}
f(t,N,\boldsymbol{A},\boldsymbol{\phi},\beta) =  \sum\limits_{n=0}^{N} A_n \sin (n \beta t + \phi_n), \quad  {\forall t \in \mathbb{R}},
\label{eq:PFS}
\end{equation}
\noindent
where $t$ denotes the time, $N$ is the number of frequencies, $\beta$ is the angular velocity, and $\boldsymbol{A}=[A_0,A_1,...A_n]$ and \mbox{$\boldsymbol{\phi}=[\phi_0,\phi_1,...,\phi_n]$} represent the amplitude and phase of the $n^{th}$ term, respectively. In our target framework, four PFS oscillators are used to generate feet trajectories ($X,Y,Z,\theta$) in Cartesian space, and one is used to generate the arm's trajectory~($A$) in the $x$-axis direction.  We considered that $c_x = f(t,N,\boldsymbol{B},\boldsymbol{\phi},\beta)$, and by plugging it into~(\ref{eq:LIPM}), the ZMP trajectory is obtained: 
\begin{figure}[!t]
    \centering
    \begin{tabular}{c c}
        \includegraphics[width = 0.45\columnwidth]{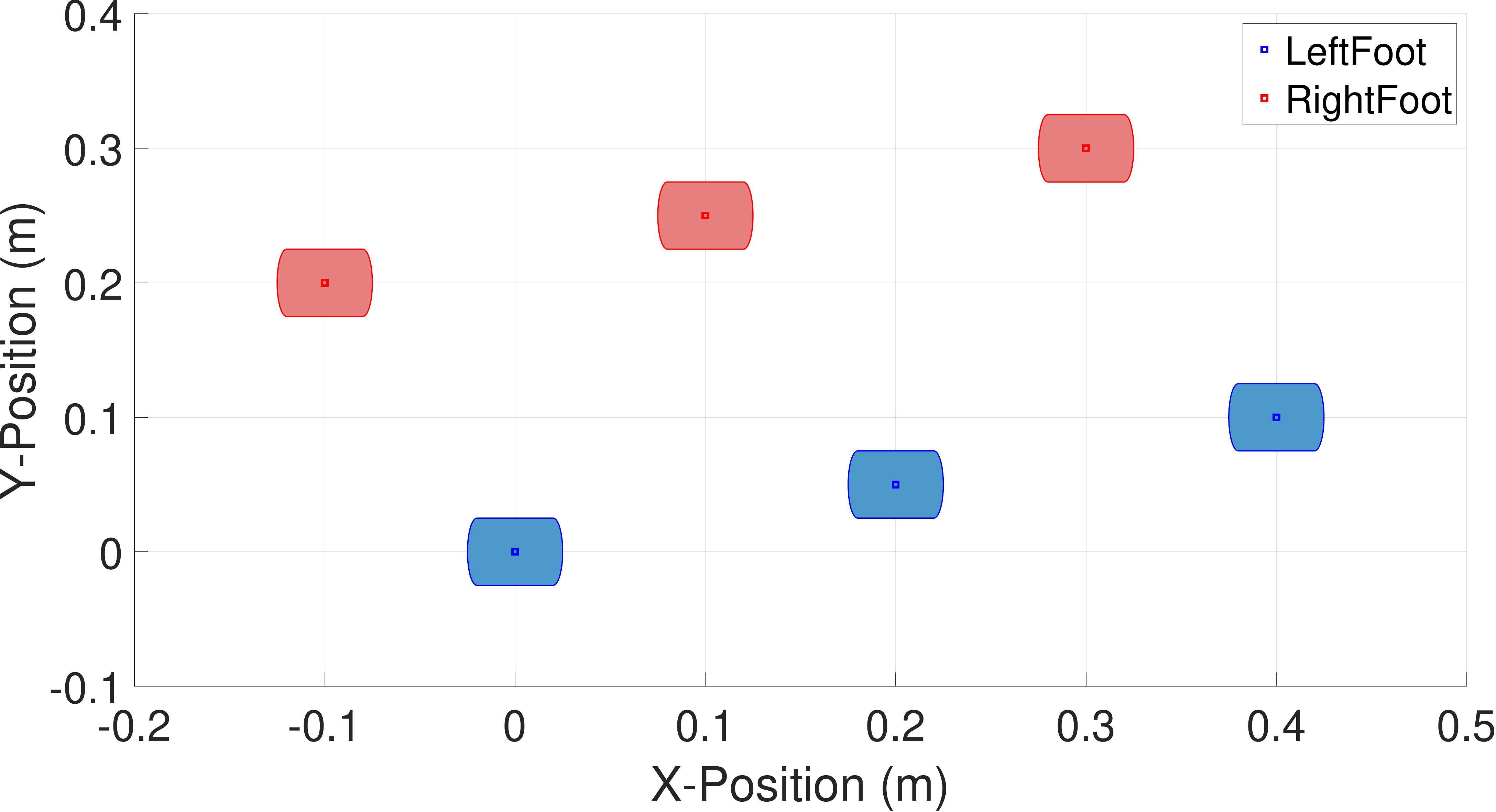}& 
        \includegraphics[width = 0.45\columnwidth]{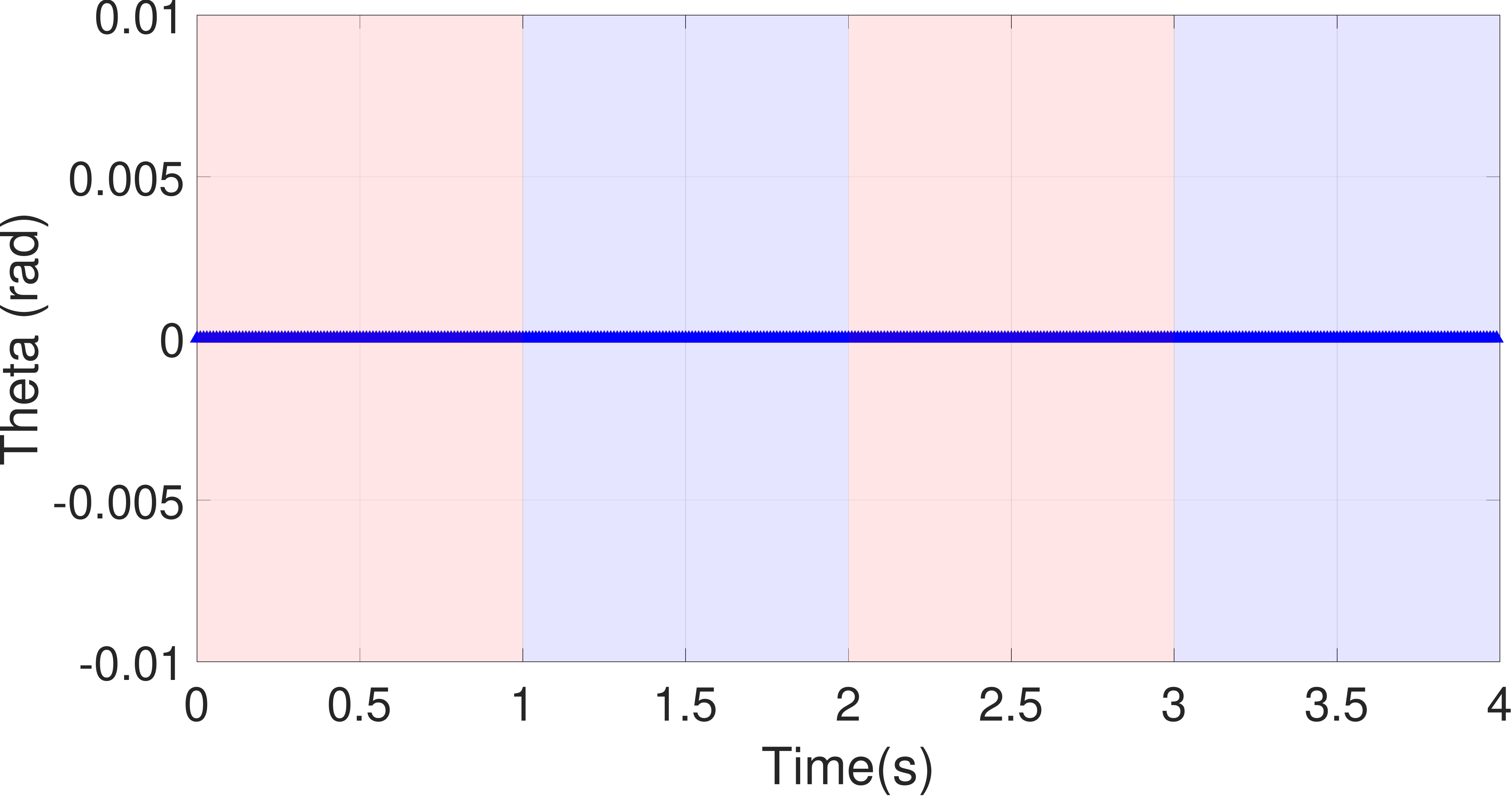}\\
        \includegraphics[width = 0.45\columnwidth]{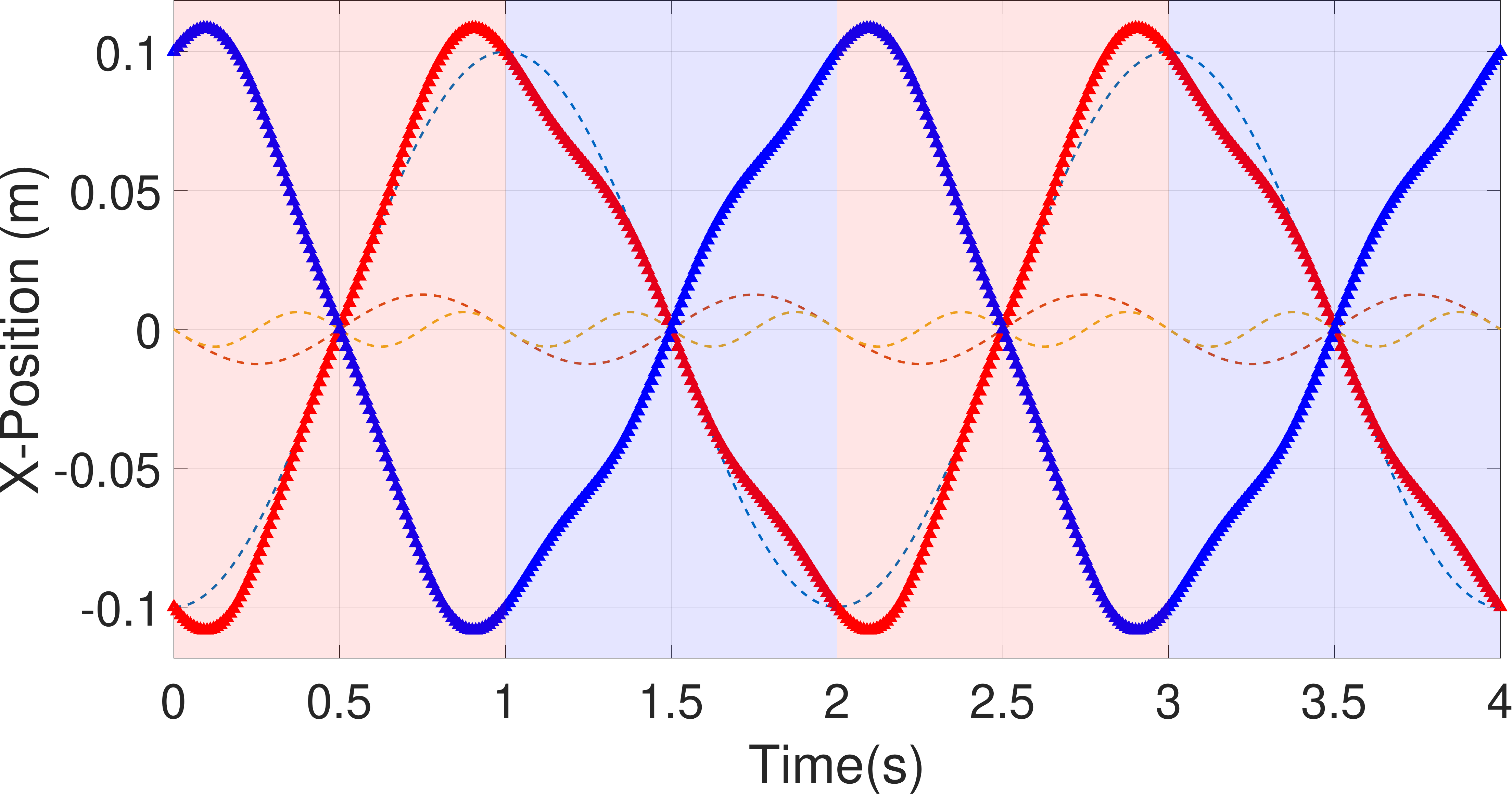}& 
        \includegraphics[width = 0.45\columnwidth]{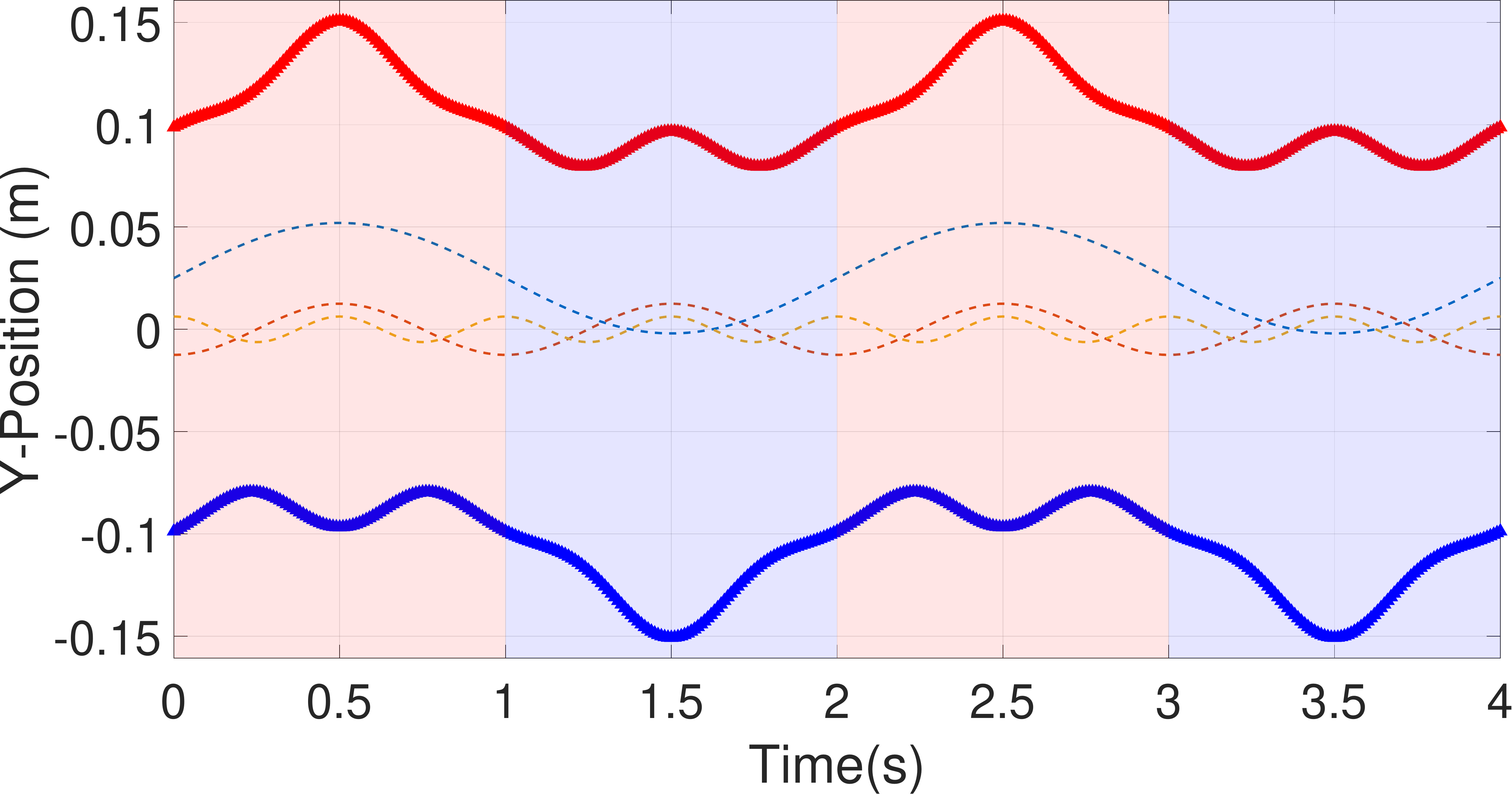}\\
         \includegraphics[width = 0.45\columnwidth]{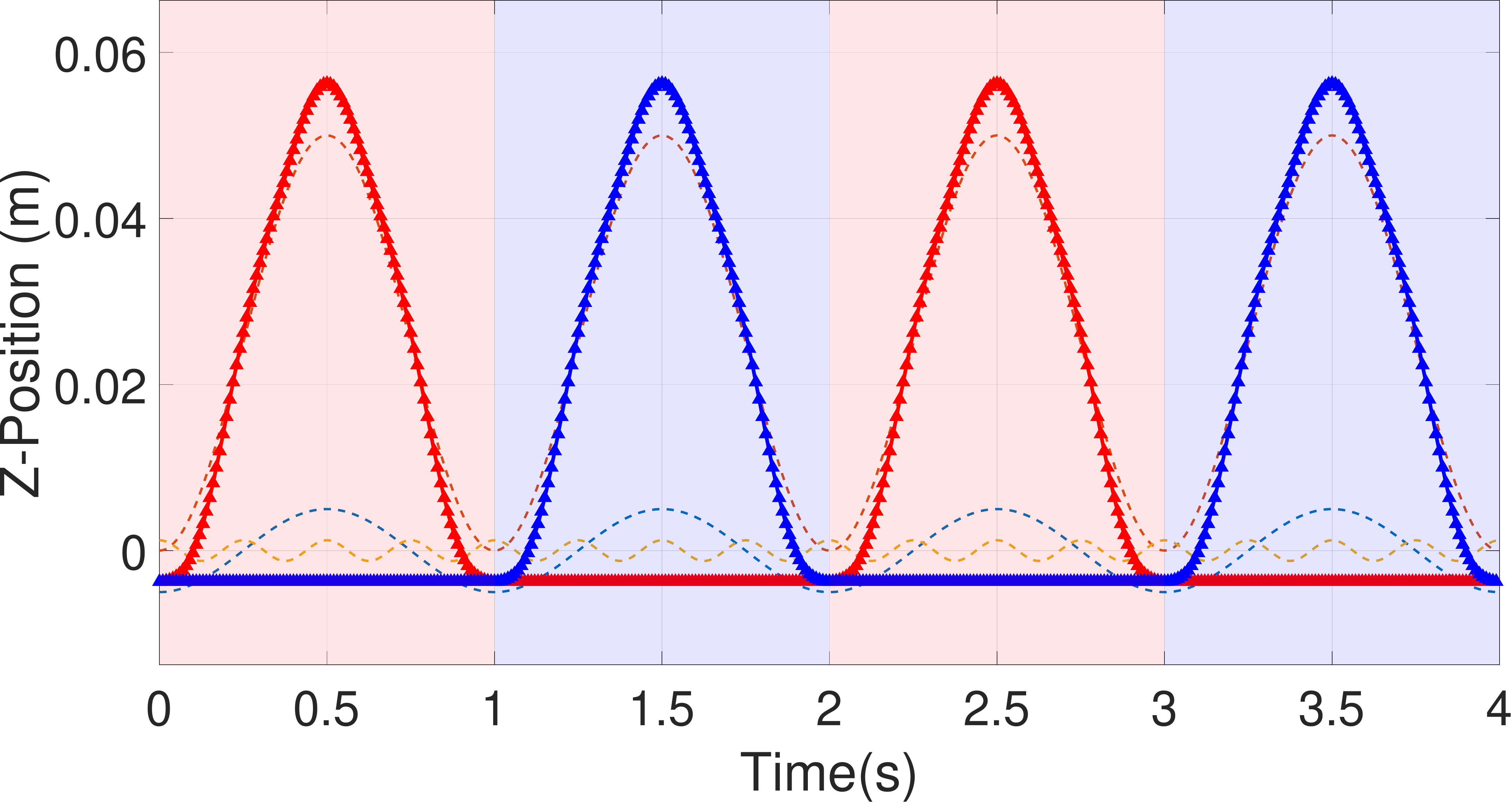}& 
        \includegraphics[width = 0.45\columnwidth]{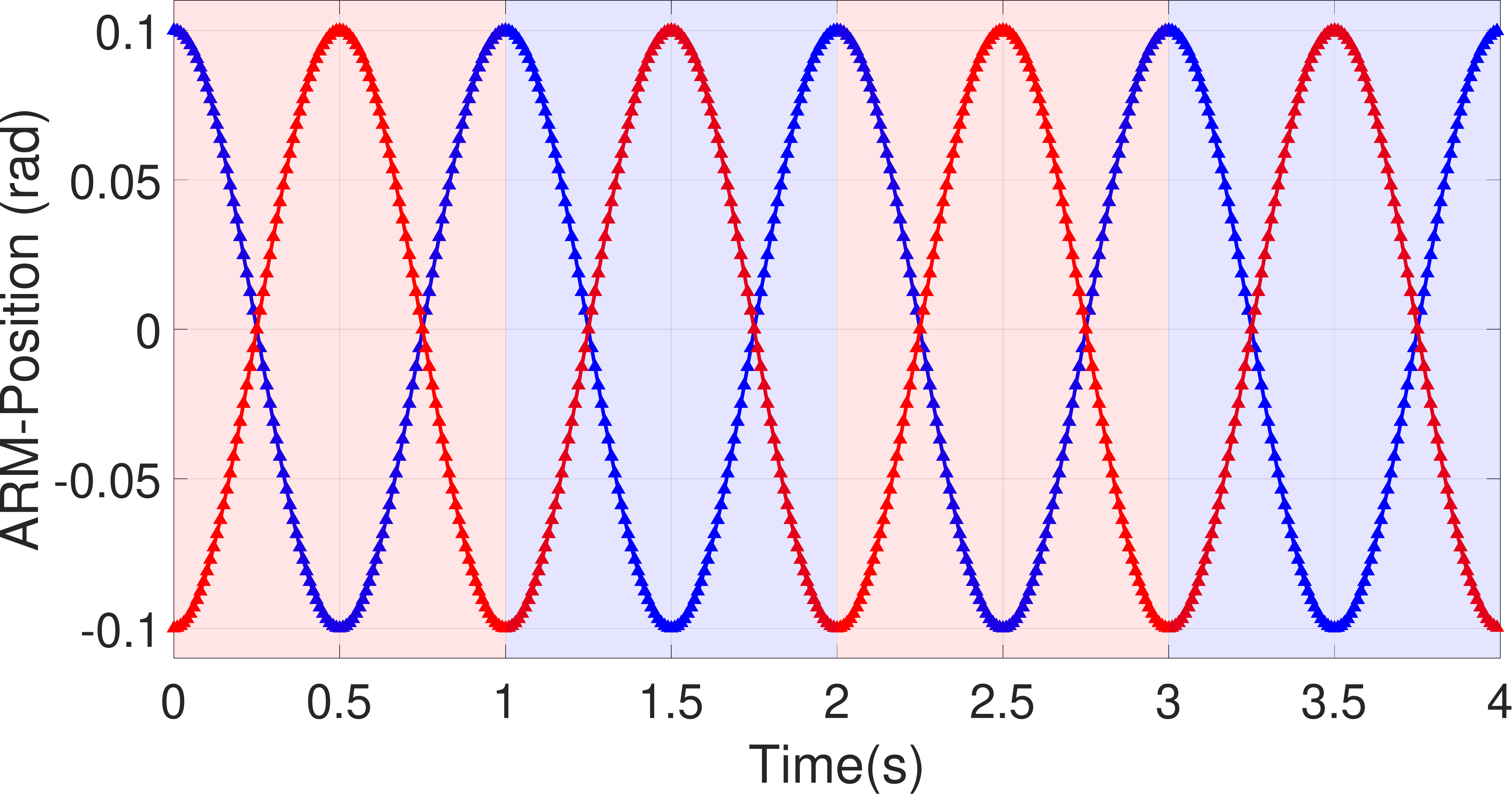}\\

    \end{tabular}
	\caption{Example trajectories for a four-steps walking \mbox{$v_x=0.2$~m/s}, \mbox{$v_y=0.05$~m/s} and \mbox{$v_\theta=0.0$~rad/s}. Dashed lines are the terms of each oscillator and thick red and blue lines represent the output of the oscillators. The red and blue phases denote when the right or left foot is moving, respectively.}
	\label{fig:traj}
\end{figure}
\begin{equation}
p_x = \sum\limits_{n=0}^{N} B_n(1 + (\frac{ n \beta }{\omega_0})^2 ) \sin (n \beta t+\phi_n) .
\label{eq:PFS_LIPM}
\end{equation}
\noindent
This equation is the core of our walk engine and shows that the COM and ZMP have the same frequencies and phase shift, but different amplitudes. This information gives us a clue to configure the oscillators and tune their parameters. The first parameter that should be configured is the number of frequencies, which is generally selected based on the oscillator's objective. For instance, the $Z$ trajectory can be generated just by one term, but another term can be added to absorb the shock during the landing phase of the swing foot. After determining the number of frequencies for each oscillator, \eqref{eq:PFS_LIPM} along with the planned foot steps are used to determine the remaining parameters, by solving nonlinear curve-fitting problems using least-squares. Fig.~\ref{fig:traj} shows the outputs of the oscillators for an example walking scenario (\mbox{$v_x=0.2$~m/s}, \mbox{$v_y=0.05$~m/s} and \mbox{$v_\theta=0.0$~rad/s)}.

	

\subsection{Walking Stabilizer}
The walking trajectories generated by the presented feed-forward walk engine do not take into consideration the state of the robot and the environment, not being robust enough to cancel the effect of uncertainties (e.g. external disturbances). To deal with this problem, we designed a set of PD controllers to stabilize the robot, based on the state of its torso, obtained through an inertial sensor. These controllers stabilize the robot 
by adjusting the torso, the ankles, the arms, and the hip joints, and can be formulated as follows:
\begin{equation}
\dot{\boldsymbol{\Phi}} - \dot{\boldsymbol{\Phi}}_d = -\boldsymbol{K}_{\Phi} (\boldsymbol{\Phi} - \boldsymbol{\Phi}_d) , 
\label{eq:PDTorso}    
\end{equation}
\noindent
where $\boldsymbol{\Phi} = [\Phi_{roll}\quad \Phi_{pitch}]^\top$ and $\dot{\boldsymbol{\Phi}}$ are the orientation and the angular velocity of the torso, $\boldsymbol{\Phi}_d$ represents the desired state and $\boldsymbol{K}_\Phi$ is the controller gain that should be tuned by an expert.

\section{Symmetry Loss Function} \label{sec:proxSymLoss}

A symmetry loss function computes a value that we seek to minimize in order to obtain a symmetric behavior. Formally, if the robot performs action $a \in \mathcal{A}_s$ in state $s \in \mathcal{S}$, it should also perform the symmetric action $g_s(a)$ in the symmetric state $f(s)$. Since symmetry is not always the main objective while learning a new behavior, a good symmetry objective function should allow asymmetric exploration in favor of a better policy.

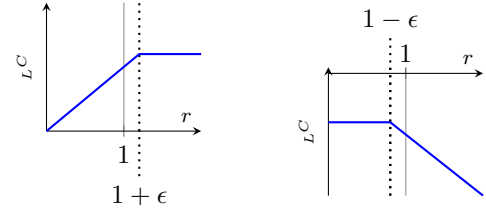
\begin{figure}[!t]
  \centering

\begin{tikzpicture}[baseline=(current bounding box.north),
  declare function={
    func(\x)= (\x < 1.2) * (x)   +
              (\x >= 1.2) * (1.2) 
   ;
  }
]
\begin{axis}[scale=0.30,
  axis x line=middle, axis y line=left,
  ymin=0, ymax=2, ytick=\empty, ylabel={\tiny$L^{C}$},
  xmin=0, xmax=2, xtick={1}, xlabel={\footnotesize$r$},
  domain=0:2,samples=6, 
  extra x ticks = {1.2},
  extra x tick labels={{$1 + \epsilon$},},
  grid=major,
      extra tick style={
        tick align=outside,
        tick pos=left,
        dotted,
        major tick length=6mm,
        line width=0.8,
        tick style={black, thick },
        grid style={black, thick}
    },
    extra x tick style={
        tick label style={yshift=0mm}
    },
    grid style={line width=0.6, draw=gray!50},
]


\addplot [blue,thick] {func(x)};

\end{axis}
\end{tikzpicture} 
\quad\quad\quad
\begin{tikzpicture}[baseline=(current bounding box.north),
  declare function={
    func(\x)= (\x < 0.8) * (-0.8)   +
              (\x >= 0.8) * (-x) 
   ;
  }
]
\begin{axis}[scale=0.30,
  axis x line=middle, axis y line=left,
  ymin=-2, ymax=0.1, ytick=\empty, ylabel={\tiny$L^{C}$},
  xmin=0, xmax=2, xtick={1}, xlabel={\footnotesize$r$},
  ylabel style={anchor=south},
  domain=0:2,samples=6, 
  extra x ticks = {0.8},
  extra x tick labels={{$1 - \epsilon$},},
  grid=major,
      extra tick style={
        tick align=inside,
        tick pos=left,
        dotted,
        x tick label style={above,yshift=3.5mm},
        major tick length=21mm,
        line width=0.8,
        tick style={black, thick },
        grid style={black, thick}
    },
    tick label style={above,yshift=1mm},
    grid style={line width=0.6, draw=gray!50},
]


\addplot [blue,thick] {func(x)};

\end{axis}
\end{tikzpicture}

\caption{Plots for PPO's surrogate objective function $L^{C}$ as a function of ratio $r$, for a single time step, for a positive advantage estimate (on the left) or a negative advantage estimate (on the right).}
\label{fig:ppo_plots}
    
\end{figure}

To understand the effect of the proposed loss function, we must analyze the most common implementation of PPO, where the policy and value networks do not share parameters. The PPO's objective $L^{PPO}$ can then be expressed as:

\begin{align}
    L^{PPO}(\theta, \omega)&=\hat{\mathbb{E}}_t \left[\; L_{t}^{C}(\theta) - L_{t}^{VF}(\omega) + cH(\theta) \;\right], \\
    \text{with} \quad
    L_{t}^{C}(\theta)&=\min\left(r_t(\theta)\hat{A}_t \;,\; (1+\sgn(\hat{A}_t)\epsilon)\hat{A}_t\right), \label{eq:lclip}\\
    \text{and} \quad\;\;\; r_t(\theta)&=\frac{\pi_{\theta}(a_t \mid s_t)}{\pi_{\theta_{old}}(a_t \mid s_t)},\label{eq:ppo_ratio}
\end{align}
\noindent
where the stochastic policy $\pi_\theta$ is parameterized by $\theta$ and the value function by $\omega$. $\pi_{\theta_{old}}$ is a copy of the policy before each update, $\hat{A}_t$ is the estimator of the advantage function, $L_{t}^{VF}$ is a squared error loss to update the value function, $\epsilon$ is a clipping parameter, $c$ is a coefficient and $H$ is the policy's distribution entropy. The expectation $\hat{\mathbb{E}}_t$ indicates the empirical average over a finite batch of samples. The main objective of this algorithm is to keep the policy update within a trust region, preventing greedy updates that can be detrimental to learning. This behavior is formalized in the surrogate objective $L_{t}^{C}$, as depicted in Fig.~\ref{fig:ppo_plots} for a single time step. 

Since the gradient $\nabla L^{PPO}(\theta, \omega)$ with respect to $\theta$ ignores $L_{t}^{VF}(\omega)$, and $c$ can typically be zeroed without negative impact, it is enough to analyze the merge of \eqref{eq:lclip} with an arbitrary symmetry loss $L_t^S$:

\begin{equation}
    \label{eq:ppo+s}
    L^{PPO+S}(\theta)=\hat{\mathbb{E}}_t \left[\; L_{t}^{C}(\theta)  - L_t^S(\theta)\;\right].
\end{equation}

Within $L_{t}^{C}$, the value of the ratio $r_t$ before the update is $1$. During the update, considering all the mini-batches and epochs before switching again to the sample acquisition stage, the ratio tends to remain near $1$, or at least in the same order of magnitude. The advantage estimate $\hat{A}_t$, under the most popular implementations of PPO \cite{openAIbaselines,stable_baselines}, has a mean of 0, because each advantage estimates batch is z-score normalized. However, when computing the gradient, since we are considering only a mini-batch, the mean is not exactly zero. In practice, for the problem formalized in this work, $L_{t}^{C}$ ranges from $-1.8\times 10^{-2}$ to $-1.0\times 10^{-2}$. It is important to retain that this value has a low order of magnitude and is relatively stable. 

\subsection{Mirror Symmetry Loss}

The mirror symmetry loss function proposed by Yu et al. \cite{yu2018learning} \mbox{$w \cdot \sum_i\|(\pi_\theta(s_i)-g_{f(s_i)}(\pi_\theta(f(s_i)))\|^2$}, where $w$ is a weight factor, computes the square error between the mean action taken by the stochastic policy before and after the symmetric transformation. On the one hand, this means that after crossing a certain asymmetry threshold, the loss dominates the policy gradient. On the other hand, if that threshold is crossed to the symmetric side, the symmetry loss loses influence in the policy update. Therefore, $w$ does not dictate the weight of the symmetry loss in a consistent way, when computing the gradient, but rather the position of the symmetry threshold. Moreover, since the square error is based on actions, instead of probabilities, the weight of the symmetry loss is also dependent on the action space.

\subsection{Proximal Symmetry Loss}

To harmonize the effect of the symmetry loss, we propose the proximal symmetry loss function

\begin{equation}
    \label{eq:psl}
    L^{PSL}(\theta)=-w \cdot \hat{\mathbb{E}}_t \left[\; \min(x_t(\theta),1+\epsilon)\;\right],
\end{equation}

where $w$ is a weight factor, $\epsilon$ is a clipping parameter shared with $L^{PPO}$, and $x_t(\theta)$ is a symmetry probability ratio. The idea behind this equation is to leverage the trust region concept that PPO (Proximal Policy Optimization) uses to iteratively reduce the model's asymmetry, hence the name "Proximal Symmetry Loss". In this sense, $L^{PSL}$ is comparable to \eqref{eq:lclip}, with the trust region defined by the same clipping parameter $\epsilon$, although there are two main differences. First, \eqref{eq:psl} does not need an advantage estimator to define the direction of the update, since the objective of this loss is to always increase or maintain symmetry. Note that, if beneficial to the policy, the RL algorithm may still choose to decrease symmetry, as $L_{t}^{C}(\theta)$ may be prioritized over $L_t^S(\theta)$ in \eqref{eq:ppo+s}. Secondly, the symmetry probability ratio is analogous to \eqref{eq:ppo_ratio} but there are some distinctions:

\begin{align}
    x_t(\theta)&= \frac{\min(\pi_\theta(a_t' \mid f(s_t)), \pi_{\theta_{old}}(a_t \mid s_t))}{\pi_{\theta_{old}}(a_t' \mid f(s_t))} , \label{eq:sym_ratio}\\
    \text{with} \quad\quad\;\; a_t'&= g_{s_t}(\pi_{\theta_{old}}(s_t \mid \sigma=0)),\label{eq:sym_a_mean}
\end{align}

\noindent
where $\pi_{\theta}(\cdot \mid s)$ and $\pi_{\theta_{old}}(\cdot \mid s)$ represent probability distributions during, and before the update, respectively. $\pi_{\theta_{old}}(s \mid \sigma=0)$ denotes the deterministic action chosen by the policy for state $s$, before the update, if the standard deviation $\sigma$ of the stochastic policy's distribution was zero. In other words, the resulting action is given by the mean of the policy's distribution. The only term that is not constant during a policy update is $\pi_\theta(a_t' \mid f(s_t))$, and only $\theta$ is being optimized. Note that $\theta_{old}$ is constant because it represents the policy's parameters before the update.

To provide an intuitive view of $x(\theta)$ for a single time step, consider a simplified function $x'(\theta)$, where $\pi_{\theta_{old}}(a \mid s)$ is ignored, such that $\mbox{$x'(\theta)=\frac{\pi_\theta(a' \mid f(s))}{\pi_{\theta_{old}}(a' \mid f(s))}$}$. Additionally, hereinafter, when referring to the relative conditional probability of the sampled action $a$ or the symmetric action $a'$, the precondition states $s$ and $f(s)$, respectively, are implicit. Ultimately, the role of $x'(\theta)$ is to increase the relative probability of the symmetric action $a'$. In comparison with \eqref{eq:ppo_ratio} there are 3 important distinctions:

\begin{itemize}
    \item The symmetric state $f(s)$ may never occur in the learning process. In fact, considering a continuous state space, the probability of $f(s)$ ever occurring during training is zero.
    
    \item The symmetric action $a'$ is not obtained from the sampled action $a$. During training, the PPO algorithm runs the stochastic policy $\pi_{\theta_{old}}$ in the provided environment to create a batch of data. This means that every action was sampled from a probability distribution with non-zero variance. Computing the symmetric action based on these samples would not only add unnecessary variance to the symmetry loss, but the final policy updates would also produce a bias in the final deterministic policy. Therefore, the artificial samples $a_t'$ are created by applying a symmetric transformation to the mean of probability distribution $\pi_{\theta_{old}}$, given the sampled states $s_t$. Note that this does not affect exploration. If, for a given time step in the future, state $f(s)$ is reached, where $s$ is the current state, the action would still be stochastic.
    
    \item The restriction to stop motivating a probability increase is based on symmetry. PPO's ratio $r_t(\theta)$ was naturally limited by the advantage estimate, i.e., if the advantage estimate was positive, the gradient would incentivize a probability increase for a certain action in a given state, and the opposite would happen for a negative estimate. Similarly, $x(\theta)$ needs to stop motivating a probability increase, when the symmetric action is as likely or more likely than the sampled action. This requirement explains the introduction of $\pi_{\theta_{old}}(a_t \mid s_t)$ in \eqref{eq:sym_ratio}. Consider a new ratio $\mbox{$z_t(\theta)=\frac{\pi_{\theta_{old}}(a_t \mid s_t)}{\pi_{\theta_{old}}(a_t' \mid f(s_t))}$}$ which is positive when the sampled action is more likely than the symmetric action, given states $s_t$ and $f(s_t)$, and negative otherwise. Additionally, note that \eqref{eq:sym_ratio} can be rewritten as $x_t(\theta)=min(x_t'(\theta),z_t(\theta))$. In conclusion, the gradient of $x_t(\theta)$ is positive as long as $x_t' < z_t$. The impact of this behavior in the proximal symmetry loss will be examined later. 
    
\end{itemize}

At this point, it is possible to examine how the proximal symmetry loss behaves in different circumstances. Fig.~\ref{fig:psl_plots} describes this behavior for a single time step, where $z$ can be in one of 3 possible groups:

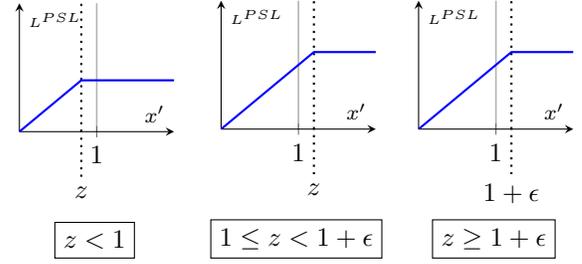
\begin{figure}[!t]
  \centering
  \begin{tikzpicture}[
  declare function={
    func(\x)= (\x < 0.8) * (x)   +
              (\x >= 0.8) * (0.8) 
   ;
  }
]
\begin{axis}[scale=0.30,
  axis x line=middle, axis y line=middle,
  ymin=0, ymax=2, ytick=\empty, ylabel={\tiny$L^{PSL}$},
  xmin=0, xmax=2, xtick={1}, xlabel={\footnotesize$x'$},
  domain=0:2,samples=6, 
  extra x ticks = {0.8},
  extra x tick labels={{$z$},},
  grid=major,
      extra tick style={
        tick align=outside,
        tick pos=left,
        dotted,
        major tick length=6mm,
        line width=0.8,
        tick style={black, thick },
        grid style={black, thick}
    },
    extra x tick style={
        tick label style={yshift=0mm}
    },
    grid style={line width=0.6, draw=gray!50},
]


\addplot [blue,thick] {func(x)};

\end{axis}
\node[below, draw] at (1,-1.2) {$z<1$};
\end{tikzpicture}
\quad
\begin{tikzpicture}[
  declare function={
    func(\x)= (\x < 1.2) * (x)   +
              (\x >= 1.2) * (1.2) 
   ;
  }
]
\begin{axis}[scale=0.30,
  axis x line=middle, axis y line=middle,
  ymin=0, ymax=2, ytick=\empty, ylabel={\tiny$L^{PSL}$},
  xmin=0, xmax=2, xtick={1}, xlabel={\footnotesize$x'$},
  domain=0:2,samples=6, 
  extra x ticks = {1.2},
  extra x tick labels={{$z$},},
  grid=major,
      extra tick style={
        tick align=outside,
        tick pos=left,
        dotted,
        major tick length=6mm,
        line width=0.8,
        tick style={black, thick },
        grid style={black, thick}
    },
    extra x tick style={
        tick label style={yshift=0mm}
    },
    grid style={line width=0.6, draw=gray!50},
]


\addplot [blue,thick] {func(x)};

\end{axis}
\node[below, draw] at (1,-1.2) {$1\leq z < 1+\epsilon$};
\end{tikzpicture} 
\quad
\begin{tikzpicture}[
  declare function={
    func(\x)= (\x < 1.2) * (x)   +
              (\x >= 1.2) * (1.2) 
   ;
  }
]
\begin{axis}[scale=0.30,
  axis x line=middle, axis y line=middle,
  ymin=0, ymax=2, ytick=\empty, ylabel={\tiny$L^{PSL}$},
  xmin=0, xmax=2, xtick={1}, xlabel={\footnotesize$x'$},
  domain=0:2,samples=6, 
  extra x ticks = {1.2},
  extra x tick labels={{$1+\epsilon$},},
  grid=major,
      extra tick style={
        tick align=outside,
        tick pos=left,
        dotted,
        major tick length=6mm,
        line width=0.8,
        tick style={black, thick },
        grid style={black, thick}
    },
    extra x tick style={
        tick label style={yshift=0mm}
    },
    grid style={line width=0.6, draw=gray!50},
]


\addplot [blue,thick] {func(x)};

\end{axis}
\node[below, draw] at (1,-1.2) {$z\geq1+\epsilon$};
\end{tikzpicture} 

\caption{Plots for the proximal symmetry loss as a function of ratio $x'=\pi_\theta(a' \mid f(s))\;/\;\pi_{\theta_{old}}(a' \mid f(s))$ for one time step. From left to right, each plot indicates the loss behavior when $z<1$, $1\leq z < 1+\epsilon$, and $z\geq 1 + \epsilon$, respectively, where $z=\pi_{\theta_{old}}(a \mid s)\;/\;\pi_{\theta_{old}}(a' \mid f(s))$.}
\label{fig:psl_plots}
    
\end{figure}

\begin{itemize}
    \item The most common scenario is when $z\geq 1+\epsilon$ (see Fig.~\ref{fig:psl_plots}, right plot), since the sampled action is usually more likely than the symmetric action. In this case, the behavior of $L^{PSL}$ is analogous to the behavior of PPO's surrogate objective function $L^C$ (see Fig.~\ref{fig:ppo_plots}, left plot). As in the original algorithm, the idea is to keep the symmetry update within a trust region, preventing greedy updates that can be detrimental to learning. Policy updates can still exceed the trust region, as there is no hard limit, but there is no explicit incentive for that outcome. Also, the gradient is constant, whether the relative probability of the symmetric action is high or low, i.e., whether the actual policy is currently symmetric or not. In this way, the influence of the proposed loss is harmonized with PPO, and $w$ carries an intuitive meaning.
    \item A less common scenario is when $z<1+\epsilon$. In this case, if $z<1$ (see Fig.~\ref{fig:psl_plots}, left plot), the symmetric action is more likely than the sampled action, and there is no incentive to change it, unless this relation changes.
    \item If $1\leq z < 1+\epsilon$ (see Fig.~\ref{fig:psl_plots}, middle plot), there is still an incentive to increase the relative probability of the symmetric action, but the trust region is smaller. 
\end{itemize}

However, the described behavior raises another question: why is the relative probability of the symmetric action never decreased? Assuming that policy $\pi_\theta$ follows a normal distribution with a reasonable variance, it is safe to assume that when dealing with asymmetric behaviors, $z$ can easily reach values with an order of magnitude of 100 or more. In practice, this means that the symmetric action will probably never be chosen in the near future, and PPO will not have a chance to evaluate it. Therefore, it makes sense to increase the relative probability of the symmetric action until it is at least as likely as the sampled action. When this happens, the action space region in the vicinity of the symmetric action has a plausible chance of being chosen by the policy, and be evaluated by PPO. In conclusion, not deliberately decreasing the probability of the symmetric action is a design choice to favor asymmetric exploration over perfect symmetry, unless the user has explicitly considered symmetry in the reward function. This decision simplifies the proximal symmetry loss and reduces its interference in training when the policy is already symmetric.

Within $x_t(\theta)$, the term $\pi_{\theta_{old}}(a_t \mid s_t)$ plays a similar role to the implicit symmetry threshold in the loss proposed by Yu et al. \cite{yu2018learning}. The difference is that the threshold is now dynamic, as it depends on the standard deviation $\sigma$ of the stochastic policy $\pi_\theta$. Without this restriction, the symmetry loss would motivate the reduction of $\sigma$ to the point where exploration is halted, and numerical instability arises in the probability computations. This approach guarantees that the loss will not work against PPO, i.e., it will converge at a stable rate, dictated by $w$, whether the policy has an initially strong asymmetrical bias or not. After that bias is reduced to a certain point, the symmetry loss will reduce its influence, allowing exploration, and promoting algorithmic stability.

\subsection{Generalized Proximal Symmetry Loss}

The proposed proximal symmetry loss function can be extended to an arbitrary number of symmetries with minimal changes. Equations \eqref{eq:psl}, \eqref{eq:sym_ratio} and \eqref{eq:sym_a_mean} can be rewritten as:

\begin{align}
    L^{PSL}(\theta)&=-w \cdot \hat{\mathbb{E}}_{t,i} \left[\; \min(x_{t,i}(\theta),1+\epsilon)\;\right],
    \\[3pt]
    \text{with} \quad
    x_{t,i}(\theta)&= \frac{\min\left(\pi_\theta(a_{t,i}' \mid f_i(s_t)), \pi_{\theta_{old}}(a_t \mid s_t)\right)}{\pi_{\theta_{old}}(a_{t,i}' \mid f_i(s_t))} , 
    \\[3pt]
    \text{and} \quad\quad\;\; a_{t,i}'&= g_{s_t, i}\left(\pi_{\theta_{old}}(s_t \mid \sigma=0)\right),
\end{align}

\noindent
where the expectation $\hat{\mathbb{E}}_{t,i}$ indicates the empirical average over a finite batch of samples, such that each sample is characterized by a total of $N$ symmetries, and $i \in \{1,2,3,...,N\}$. Likewise, functions $f_i(s)$ and $g_{s,i}(a)$ apply symmetry transformation $i$ to state $s$, and $s$-dependent action $a$.

\section{Simulated Scenarios}
\begin{figure}[!t]
    \centering
    \includegraphics[width = 0.96\columnwidth]{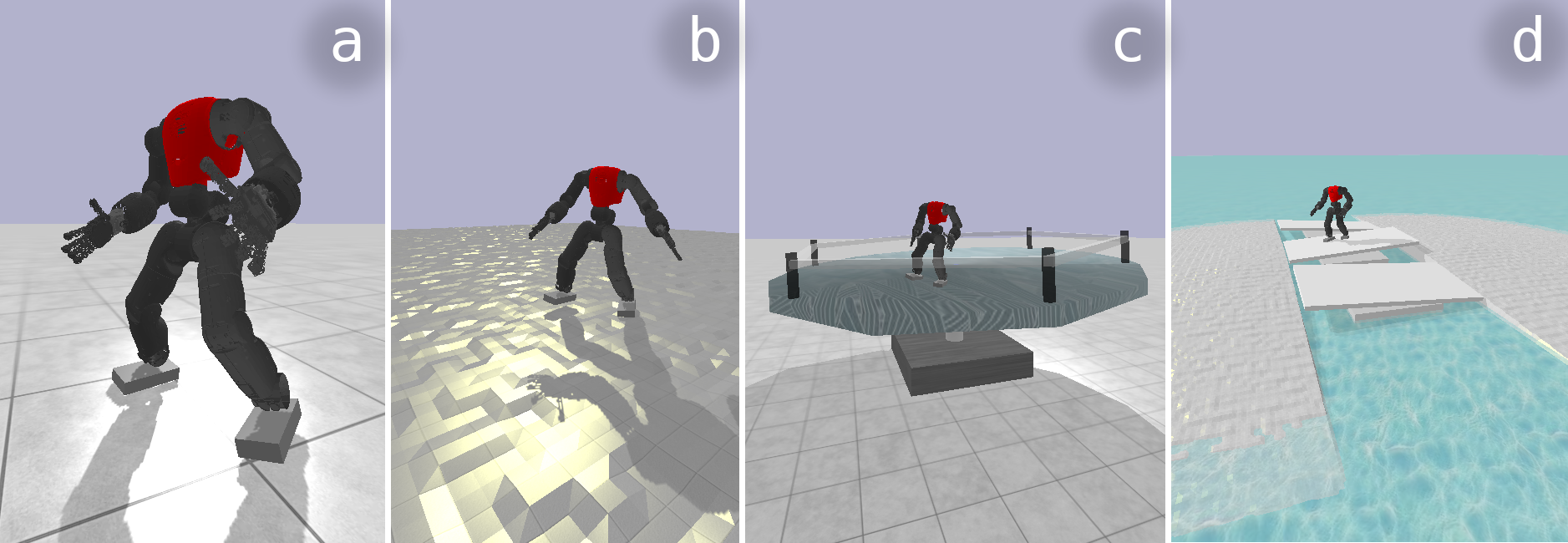}
	\caption{Simulation scenarios designed in PyBullet \cite{coumans2020}: flat surface (\textbf{a}), uneven terrain (\textbf{b}), autonomous tilting platform (\textbf{c}), and island scenario (\textbf{d}) which combines the challenges of the previous ones. The robot was trained on scenario (\textbf{b}), and the others were used to test the framework in unforeseen circumstances.}
	\label{fig:scenarios}
\end{figure}

Four scenarios were designed to test the walking capabilities of the COMAN humanoid robot in harsh conditions (see Fig.~\ref{fig:scenarios}). They are ordered by increasing difficulty: flat surface (\textbf{a}); uneven terrain with 2~cm perturbations (\textbf{b}); tilting platform that rotates autonomously, following a random pattern with a 4.5-degree maximum tilt (\textbf{c}); two islands scenario that combines uneven terrain (1~cm perturbations), a constant 3-degree slope, and unstable seesaws between both islands with a 4-degree maximum tilt (\textbf{d}). 

The agent was optimized in scenario (\textbf{b}) without prior knowledge of the terrain. The optimization encompassed all the outputs of the \texttt{residuals} block in Fig.~\ref{fig:arch}: target joint positions, step size $w_x$, $w_y$ and rotation $w_\theta$, target COM, step duration and PD gains. The other three scenarios were designed to assess the ability of the framework to generalize to new conditions, and compare its raw performance against a baseline. 

The robot is initially placed in the center of the map, except in (\textbf{d}), where it is placed on the left island. Then, the high-level controller (see Fig.~\ref{fig:arch}) generates a new objective, composed of a 2D position $(x,y)$ and a desired final orientation $\theta$. Each new objective is generated randomly, such that the new position is no farther than 5~m from the robot's current position. In scenario (\textbf{c}), this position is restricted to the platform's surface, while in (\textbf{d}), it is restricted to the two islands and three seesaws. Additionally, in the latter scenario, every position in a straight line that connects two consecutive objectives must follow the same restrictions. The idea is to allow the robot to use the shortest path to the next objective without falling to the water. The agent is rewarded when the robot reduces its distance to the objective and vice versa:

\begin{equation}
    R_t = k_1(d_{t-1}-d_t) + k_2(\alpha_{t-1}-\alpha_t),
\end{equation}
\noindent
where $d_t$ is the linear distance between robot and objective, in meters, $\alpha_t$ is the angular distance, in radians, and $k_1=2$ and $k_2=3.5$ are proportional constants defined empirically. Each objective has a conquer time of 1.5~s, i.e., the robot must be within a small area (less than 15~cm from the target position) for 1.5~s for the high-level controller to generate a new objective. This approach intends to motivate a stable walking-in-place behavior. Otherwise, the robot is always rushing to the next objective, without mastering the subtler skills. An episode ends when the robot falls or after a maximum period of 1000~s, although the latter restriction was not applied during the learning stage. To get a higher reward, the agent must learn to maximize its speed, stability and accuracy in order to efficiently conquer each objective.

In the testing stage, in addition to the terrain conditions, an external force was applied to the robot at fixed intervals of 4 seconds, in a random direction parallel to the ground, with a constant duration of 25~ms. The applied force magnitude was initially set to 300~N, and it was then increased to 400~N. Moreover, a relative noise value was added to all observations, ranging from $-10\%$ to $+10\%$, i.e., each observation was multiplied by a random number between 0.9 and 1.1, sampled from a continuous uniform distribution at every time step.

\section{Experiments}

To evaluate the performance of the framework, the results were compared with a baseline, which is only composed of the walk engine presented in Section \ref{sec:walk_engine}. Table \ref{tb:results} shows a comparison between the baseline and the best model developed by the proposed framework. Both models were tested in the four scenarios presented in Fig.~\ref{fig:scenarios}, using a force magnitude of 0~N, 300~N and 400~N. Regarding the baseline, the 400~N line was omitted because the robot could not withstand a single push without losing its balance. The same logic can be applied to 300~N for scenario (\textbf{b}), (\textbf{c}) and (\textbf{d}). There are two metrics per scenario: the average episode duration $t$, measured in seconds; and the average reward per second (obtained by dividing the episode's total reward by its duration). Each simulation configuration was executed for 1000 episodes.

\begin{table}[]
    \centering
    \setlength\tabcolsep{0pt} 
    \caption{Results comparison for each simulated scenarios}
	\label{tb:results}
\begin{tabular}{@{\hskip 4pt}l@{\hskip 4pt}|P{9mm}|P{8mm}P{8mm}|P{8mm}P{8mm}|P{8mm}P{8mm}|P{8mm}P{8mm}}
\multirow{2}{*}{} & \multirow{2}{*}{Force} & \multicolumn{2}{c}{(\textbf{a})} & \multicolumn{2}{c}{(\textbf{b})} & \multicolumn{2}{c}{(\textbf{c})} & \multicolumn{2}{c}{(\textbf{d})} \\
 &  & $t$(s) & $r/t$ & $t$(s) & $r/t$ & $t$(s) & $r/t$ & $t$(s) & $r/t$ \\  \hline 
 \multirow{2}{*}{Baseline} & 0N &  158 & 0.46 & 1 & -0.11  & 2 & -0.51 & 2 & -0.30 \\
 & 300N \vspace{1mm} & 25 & 0.42 & - & - & - & - & - & - \\ 
\multirow{3}{*}{\begin{tabular}[c]{@{}l@{}}Best\\ Model\end{tabular}} & 0N &  976 & 0.94 & 520 & 0.84 & 35 & 0.64 & 55 & 0.50  \\
 & 300N & 332 & 0.90 & 97 & 0.79 & 15 & 0.25 & 29 & 0.45 \\
 & 400N & 55 & 0.86 & 29 & 0.74 & 12 & 0.28 & 20 & 0.41
\end{tabular}
\end{table}

As aforementioned, the simulation was limited to a maximum period of 1000~s during the testing stage. In scenario (\textbf{a}), the best model was able to survive, on average, for 976~s, which characterizes the corresponding gait as very stable. Without this limitation, some simulation threads can run almost indefinitely. As expected, the trained model generalized better for scenario (\textbf{a}), than (\textbf{c}) or (\textbf{d}). However, after visually inspecting the results (see simulation videos at \url{https://youtu.be/mpGYdo_PIMo}), the robot displays very interesting and human-like skills in scenarios (\textbf{c}) and (\textbf{d}). It must be noted that the best model is built on top of the baseline walk engine, which falls almost immediately in scenarios other than (\textbf{a}) and has no prior knowledge of the terrain.

After adding random external forces, the best model can sometimes recover by stepping in the direction of the push, although it was not trained to do that. We conclude that learning to walk on uneven terrain has given the robot the ability to recover its balance by stepping in some direction, even if that results in getting farther from the objective. As an example, in scenario (\textbf{b}), where the robot learned, it can resist 400~N pushes for 29 seconds, on average. This is equivalent to approximately 7 pushes per episode, on average.

\subsection{Proximal Symmetry Loss}

\begin{figure}[!t]
	\centering 
	
	\hspace{0.5mm}
	\begin{tikzpicture}[]
	
	\pgfplotsset{
		height=3.8cm, width=8.0cm, compat=1.15,
	}

	\begin{axis}[
		legend pos=north west,
		legend cell align={left},
		ymax = 1800,
		ytick distance= 400,
		xtick distance= 10,
		xticklabels={,,},
        minor x tick num=9,
        ymin=0, xmin=0, xmax=100,
        grid=both,
        minor grid style={black!5},
        major grid style={black!20},
        minor x tick style={black!10},
        legend style={font=\scriptsize},
        dottedstyle
	]
	
	\addplot[rdot] table [x expr=\coordindex/4.45, y=YU1, col sep=comma]{Plots/16k.csv};
	\addplot[rdot] table [x expr=\coordindex/4.45, y=YU2, col sep=comma]{Plots/16k.csv};
	\addplot[rdot] table [x expr=\coordindex/4.45, y=YU3, col sep=comma]{Plots/16k.csv};
	\addplot[rdot] table [x expr=\coordindex/4.45, y=YU4, col sep=comma]{Plots/16k.csv};
	
	\addplot[bdot] table [x expr=\coordindex/4.45, y=mine1, col sep=comma]{Plots/16k.csv};
	\addplot[bdot] table [x expr=\coordindex/4.45, y=mine2, col sep=comma]{Plots/16k.csv};
	\addplot[bdot] table [x expr=\coordindex/4.45, y=mine3, col sep=comma]{Plots/16k.csv};
	\addplot[bdot] table [x expr=\coordindex/4.45, y=mine4, col sep=comma]{Plots/16k.csv};
	
	\addplot[gdot] table [x expr=\coordindex/4.45, y=base1, col sep=comma]{Plots/16k.csv};
	\addplot[gdot] table [x expr=\coordindex/4.45, y=base2, col sep=comma]{Plots/16k.csv};
	\addplot[gdot] table [x expr=\coordindex/4.45, y=base3, col sep=comma]{Plots/16k.csv};
	\addplot[gdot] table [x expr=\coordindex/4.45, y=base4, col sep=comma]{Plots/16k.csv};
	
	\addplot[blin] table [x expr=\coordindex/4.45, y=avgMine, col sep=comma]{Plots/16k.csv};
	\addplot[rlin] table [x expr=\coordindex/4.45, y=avgYU, col sep=comma]{Plots/16k.csv};
	\addplot[glin] table [x expr=\coordindex/4.45, y=avgBase, col sep=comma]{Plots/16k.csv};
	
	\node[rectangle, draw=black!60, fill=white,  thick, minimum size=5mm, anchor=west, text width=4mm, align=center] at (3,400){16k};
	
	\legend{,,,,,,,,,,,,PPO+PSL,PPO+MSL,PPO}
	\end{axis}
	\end{tikzpicture}
	
	\vspace{-1mm}
	\hspace{-4mm}
	\begin{tikzpicture}[]
	
	\pgfplotsset{
		height=3.8cm, width=8.0cm, compat=1.15,
	}

	\begin{axis}[
		ylabel=Avg. Reward / Ep.,
		legend pos=north east,
		legend cell align={left},
		ymax = 1800,
		ytick distance= 400,
		xtick distance= 10,
		xticklabels={,,},
        minor x tick num=9,
        ymin=0, xmin=0, xmax=100,
        grid=both,
        minor grid style={black!5},
        major grid style={black!20},
        minor x tick style={black!10},
        dottedstyle
	]
	
    \addplot[rdot] table [x expr=\coordindex/4.45, y=YU1, col sep=comma]{Plots/8k.csv};
	\addplot[rdot] table [x expr=\coordindex/4.45, y=YU2, col sep=comma]{Plots/8k.csv};
	\addplot[rdot] table [x expr=\coordindex/4.45, y=YU3, col sep=comma]{Plots/8k.csv};
	\addplot[rdot] table [x expr=\coordindex/4.45, y=YU4, col sep=comma]{Plots/8k.csv};
	
	\addplot[bdot] table [x expr=\coordindex/4.45, y=mine1, col sep=comma]{Plots/8k.csv};
	\addplot[bdot] table [x expr=\coordindex/4.45, y=mine2, col sep=comma]{Plots/8k.csv};
	\addplot[bdot] table [x expr=\coordindex/4.45, y=mine3, col sep=comma]{Plots/8k.csv};
	\addplot[bdot] table [x expr=\coordindex/4.45, y=mine4, col sep=comma]{Plots/8k.csv};
	
	\addplot[gdot] table [x expr=\coordindex/4.45, y=base1, col sep=comma]{Plots/8k.csv};
	\addplot[gdot] table [x expr=\coordindex/4.45, y=base2, col sep=comma]{Plots/8k.csv};
	\addplot[gdot] table [x expr=\coordindex/4.45, y=base3, col sep=comma]{Plots/8k.csv};
	\addplot[gdot] table [x expr=\coordindex/4.45, y=base4, col sep=comma]{Plots/8k.csv};
	
	\addplot[rlin] table [x expr=\coordindex/4.45, y=avgYU, col sep=comma]{Plots/8k.csv};
	\addplot[blin] table [x expr=\coordindex/4.45, y=avgMine, col sep=comma]{Plots/8k.csv};
	\addplot[glin] table [x expr=\coordindex/4.45, y=avgBase, col sep=comma]{Plots/8k.csv};

    \node[rectangle, draw=black!60, fill=white,  thick, minimum size=5mm, anchor=west, text width=4mm, align=center] at (3,400){8k};
    
	\end{axis}
	\end{tikzpicture}
	
	
	\vspace{-0.5mm}
	\hspace{2.7mm}
	\begin{tikzpicture}[]
	
	\pgfplotsset{
		height=3.8cm, width=8.0cm, compat=1.15,
	}

	\begin{axis}[
		xlabel=Time steps ($\times 10^6$),
		legend pos=north east,
		legend cell align={left},
		ymax = 1800,
		ytick distance= 400,
		xtick distance= 10,
        minor x tick num=9,
        ymin=0, xmin=0, xmax=100,
        grid=both,
        minor grid style={black!5},
        major grid style={black!20},
        minor x tick style={black!10},
        dottedstyle
	]
	
    \addplot[rdot] table [x expr=\coordindex/4.45, y=YU1, col sep=comma]{Plots/4k.csv};
	\addplot[rdot] table [x expr=\coordindex/4.45, y=YU2, col sep=comma]{Plots/4k.csv};
	\addplot[rdot] table [x expr=\coordindex/4.45, y=YU3, col sep=comma]{Plots/4k.csv};
	\addplot[rdot] table [x expr=\coordindex/4.45, y=YU4, col sep=comma]{Plots/4k.csv};
	
	\addplot[bdot] table [x expr=\coordindex/4.45, y=mine1, col sep=comma]{Plots/4k.csv};
	\addplot[bdot] table [x expr=\coordindex/4.45, y=mine2, col sep=comma]{Plots/4k.csv};
	\addplot[bdot] table [x expr=\coordindex/4.45, y=mine3, col sep=comma]{Plots/4k.csv};
	\addplot[bdot] table [x expr=\coordindex/4.45, y=mine4, col sep=comma]{Plots/4k.csv};
	
	\addplot[gdot] table [x expr=\coordindex/4.45, y=base1, col sep=comma]{Plots/4k.csv};
	\addplot[gdot] table [x expr=\coordindex/4.45, y=base2, col sep=comma]{Plots/4k.csv};
	\addplot[gdot] table [x expr=\coordindex/4.45, y=base3, col sep=comma]{Plots/4k.csv};
	\addplot[gdot] table [x expr=\coordindex/4.45, y=base4, col sep=comma]{Plots/4k.csv};
	
	\addplot[rlin] table [x expr=\coordindex/4.45, y=avgYU, col sep=comma]{Plots/4k.csv};
	\addplot[blin] table [x expr=\coordindex/4.45, y=avgMine, col sep=comma]{Plots/4k.csv};
	\addplot[glin] table [x expr=\coordindex/4.45, y=avgBase, col sep=comma]{Plots/4k.csv};
	
	\node[rectangle, draw=black!60, fill=white,  thick, minimum size=5mm, anchor=west, text width=4mm, align=center] at (3,400){4k};

	\end{axis}
	\end{tikzpicture}

	\caption{Training evolution comparison between PPO, the loss function extension presented in this work (PPO+PSL), and the extension proposed by Yu et al. \cite{yu2018learning} (PPO+MSL). Each plot represents the evolution for 100 million time steps, using 8 threads and a batch size per thread of 16k, 8k and 4k, ordered from top to bottom. The dotted lines correspond to the average reward per episode for a single run. Each solid line is the average of 4 independent runs.}
	\label{fig:evolution}
\end{figure}
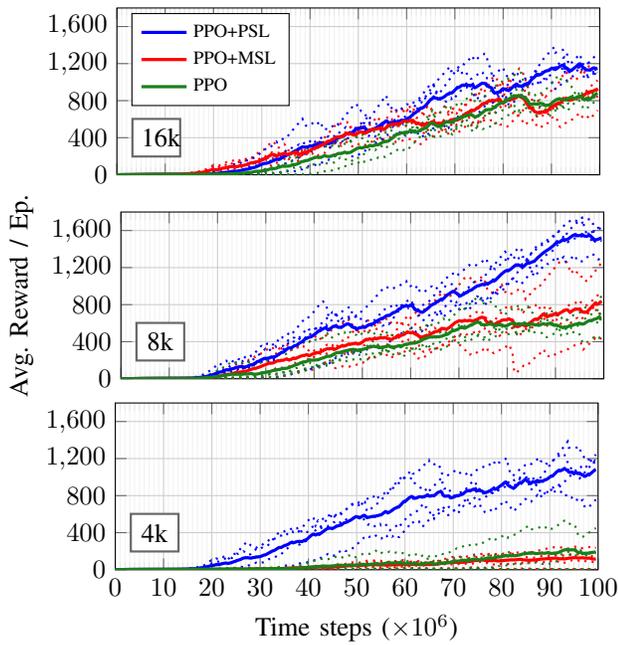

To evaluate the sample efficiency of the different symmetry loss functions, the reinforcement learning algorithm was executed with three different batch configurations, corresponding to the three plots in Fig.~\ref{fig:evolution}. All optimizations ran for 100 million time steps, using 8 threads. Analyzing the first plot, on the top, there are 12 dotted lines, which show the evolution of the average reward per episode for 12 optimization runs, with a batch size per thread of 16k samples: 4 for PPO~(green), 4 for our extension PPO+PSL (blue), and 4 for the symmetry mirror loss extension proposed by Yu et al. \cite{yu2018learning} (red). There are also 3 solid lines that follow the same color association and represent the mean of the corresponding 4 runs. The middle and bottom plots are equivalent to the top plot, except for the number of samples per batch.

This experiment confirms the intuitive notion that leveraging symmetry knowledge brings an added value to each sample. For 16k samples per batch, the performance of every approach is very similar, and above that size, it drops in the same proportion. It must be noted that the performance drops, not because of excessive samples, but because the number of updates is reduced for the same 100 million time steps. This constitutes a trade-off between number of updates and information per update. However, after a while, adding samples to a batch yields no benefit. So, each approach has a certain optimal balance. For our learning scenario, the PPO+PSL approach works best with approximately 8k samples per batch. However, the optimal batch size for PPO and PPO+PSL seems to be around 16k samples. The difference is especially evident in the bottom plot, where each batch has only 4k samples. As the optimizations converge to a global optimum, the reward grows indefinitely, since the episodes have no time limitation during learning. Therefore, this analysis is only focused on sample efficiency, not on comparing the quality of the final solution after a plateau is reached.

\section{Conclusion}

In this paper, we presented a hybrid walking framework by coupling a closed-loop CPG-ZMP walk engine with a DRL algorithm to combine the potential of both approaches. First, a feed-forward CPG-based walk engine was designed based on the LIPM and a set of PFS oscillators. Then, a walking stabilizer was formulated as a set of PD controllers, which were tuned by an expert to have a closed-loop CPG-ZMP walk engine. Additionally, we optimized a policy to adaptively update the walk engine parameters and to generate the steering commands, as well as residual joint position targets which complement the walk engine with model-free skills. This hybrid framework aims at generating robust, versatile and agile omnidirectional walking gaits by exploring the full potential of the robot, taking advantage of the analytical solution's consistency and the flexibility of residual learning. Moreover, a proximal symmetry loss function was proposed to increase the sample efficiency of PPO by leveraging MDP symmetries. To assess the performance of the overall framework, four simulation scenarios were designed. The robot was trained in only one of the scenarios. However, it was able to generalize its knowledge in unforeseen circumstances, displaying very interesting skills, which can be visualized at \url{https://youtu.be/mpGYdo_PIMo}.

\section*{Acknowledgment}
This research is supported by Portuguese National Funds through Foundation for Science and Technology (FCT) through FCT scholarship SFRH/BD/118438/2016. The second author is supported by FCT under grant SFRH/BD/139926/2018.


%




\ifCLASSOPTIONcaptionsoff
  \newpage
\fi



\bibliographystyle{IEEEtran}
\bibliography{main}
\end{document}